**Title:** The weak relationship between ankle proprioception and gait speed after stroke: a robotic assessment study

**Authors:**

Name, Highest Degree Achieved, department, institution, city, country

1. Christopher A. Johnson, MS, Henry Samueli School of Engineering Department of Biomedical Engineering, University of California, Irvine, Irvine, USA

2. Piyashi Biswas, MS, Henry Samueli School of Engineering Department of Biomedical Engineering, University of California, Irvine, Irvine, USA

3. Rubi Tapia, HS, Henry Samueli School of Engineering Department of Biomedical Engineering, University of California, Irvine, Irvine, USA

4. Jill See, DPT, Department of Rehabilitation Services, University of California at Irvine Medical Center, Orange, USA

5. Lucy Dodakian, MA, OTR/L, Department of Rehabilitation, University of California, Irvine, Irvine, USA

6. Vicky Chan, DPT, Department of Rehabilitation Services, University of California at Irvine Medical Center, Orange, USA

7. Po T. Wang, PhD, Henry Samueli School of Engineering, University of California, Irvine, Irvine, USA

8. Zoran Nenadic, D.Sc, Henry Samueli School of Engineering Department of Biomedical Engineering, University of California, Irvine, Irvine, USA

9. An H. Do, MD, School of Medicine Department of Neurology, University of California, Irvine, Irvine, USA

10. David J. Reinkensmeyer, PhD, UC Irvine School of Medicine Department of Anatomy and Neurobiology, Henry Samueli School of Engineering Department of Mechanical and Aerospace Engineering, University of California, Irvine, Irvine, USA (and affiliation 1)

Corresponding Author: Christopher Johnson, 3225 Engineering Gateway, University of California, Irvine, Irvine, CA 92697-3975, johnson4@uci.edu

**Article Word Count:** 4,205

**Number of Figures and Tables:** 5

## Abstract

**Background:** Ankle proprioceptive deficits are common after stroke and occur independently of ankle motor impairments. Despite this independence, some studies have found that ankle proprioceptive deficits predict gait function, consistent with the concept that somatosensory input plays a key role in gait control. Other studies, however, have not found a relationship, possibly because of variability in proprioception assessments. Robotic assessments of proprioception offer improved consistency and sensitivity.

**Objective:** Establish relationships between ankle proprioception, ankle motor impairment, and gait function after stroke using robotic assessments of ankle proprioception.

**Methods:** We quantified ankle proprioception using two different robotic tests (Joint Position Reproduction and Crisscross) in 39 persons in the chronic phase of stroke. We analyzed the extent to which these robotic proprioception measures predicted gait speed, measured over a long distance (6-minute walk test) and a short distance (10-meter walk test). We also studied the relationship between robotic proprioception measures and lower extremity motor impairment, quantified with measures of ankle strength, active range of motion, and the lower extremity Fugl-Meyer exam.

**Results:** Impairment in ankle proprioception was present in 87% of the participants. Ankle proprioceptive acuity measured with JPR was weakly correlated with 6MWT gait speed ($\rho = -0.34$, $p = 0.039$) but not 10mWT ($\rho = -0.29$, $p = 0.08$). Ankle proprioceptive acuity was not correlated with lower extremity motor impairment ($p > 0.2$).

**Conclusion:** These results confirm the presence of a weak relationship between ankle proprioception and gait after stroke that is independent of motor impairment.

**Keywords:** robotics, gait, proprioception, stroke rehabilitation, clinical tests, muscle strength

## Introduction

Stroke is a leading cause of disability worldwide, creates both sensory and motor deficits, and impacts the ability to complete activities of daily living [1]. There is a large body of evidence linking lower extremity (LE) motor impairment to gait function in patients with stroke [1–3]. For example, strength of the impaired ankle has often been found to be a strong predictor of gait velocity [1,3]. Somatosensory deficits of the LE are also common after stroke, and are estimated to occur in 50% of individuals [4]. The magnitude of LE sensory deficits appears to be independent of the magnitude of LE motor impairment [5–8]. However, when it comes to LE sensory deficits predicting gait function, results are less clear.

Proprioception is one's ability to integrate sensory signals to determine body and limb position, movements, and force and is important in motor planning, control, and learning [9]. Proprioception is mediated by an array of proprioceptors, including cutaneous receptors, joint mechanoreceptors, muscle spindles, and Golgi tendon organs (GTOs). Proprioceptive impairment after stroke is thought to affect the control of muscle tone, disrupt postural reflexes, and impair spatial and temporal aspects of volitional movement [10]. For the upper extremity, proprioceptive impairment also predicts the ability of persons who have experienced a stroke to benefit from rehabilitative movement training, such as constraint-induced therapy [11] or robotic hand movement training [12–14]. This suggests that proprioceptive feedback plays an important role in mediating use-dependent plasticity. For the LE, proprioceptive signals related to leg kinematics and loading are thought to play a key role in locomotor control and plasticity [15–18].

Unlike for ankle strength, however, studies have typically found no association [5–8] or only weak associations between LE sensation and gait function, quantified as gait velocity [8,19], balance ability [7,19,20], falls [7,19], and endurance [21] (Table 1). One reason may be the wide variety in methods used to quantify proprioceptive acuity [5–8]. Several studies that found no relationship have used available clinical assessments, such as the modified Nottingham Sensory Assessment [7] and the sensory scale of the Fugl Meyer Assessment [5,8], which provide basic information on an individual's ability to perceive movement and/or its direction on an ordinal scale as "absent", "impaired", or "normal" [22]. These clinical assessments of sensory impairment have limited accuracy and responsiveness [19,22]. Other assessments for LE proprioception have been developed using robotics [23,24] to address these limitations. Robotic assessments for both sensory and motor impairments are more objective (not relying as strongly on observer judgement) and accurate (e.g. able to measure exact body position/force applied), because they can deliver precise, reproducible stimuli and then measure the response to those stimuli [25].

TABLE 1

Here, we sought to clarify the relationship between ankle proprioception and gait function after stroke using robotic assessments of ankle proprioception. We used a custom-built robotic device to implement a commonly used type of proprioception test, a Joint Position Reproduction (JPR), as well as more novel proprioception assessment, Crisscross, which we recently developed to measure proprioception acuity of the fingers [26]. JPR asks individuals to copy the motion imposed on one ankle by the robotic device by actively moving their other ankle in a matching motion. Crisscross reduces the motor demand on individuals by asking them to simply push a button when they feel their ankles cross each other, as the ankles are driven through their plantar/dorsi flexion range of motion using a robotic device. Crisscross combines components of movement discrimination, movement speed, direction sense, comparison between relative limb positioning in space, and discrete position matching (where by "discrete" we mean the test subject makes one proprioceptive judgment per each movement trial). Studies that employed Crisscross with a robotic exoskeleton to measure finger proprioception have found that the test is sensitive to aging [26] and presence of a prior stroke [14], and that it predicted stroke subjects' ability to benefit from a three-week period of robotic finger training[12,13]. For our measure of gait function, we focused on standardized measures of gait speed over long and short distances.

## Methods

**Participants and Clinical Assessments**

Chronic stroke participants were enrolled in an ongoing clinical trial designed to evaluate the efficacy of a brain-computer-interface, functional electrical stimulation system for treating foot drop. The inclusion criteria were as follows: Age 18-80 years, radiologically confirmed stroke, with day of onset at least 26 weeks prior to day of randomization, Gait velocity < 0.8 m/s, foot-drop in affected limb, plantarflexor spasticity < 3 on modified Ashworth Scale, walk > 10 m (with or without ankle foot orthosis (AFO), and cane or walker permitted) at a supervised level. Only baseline measurements were used for analysis.

A trained physical therapist assessed each participant. The following measures were taken: National Institutes of Health Stroke Scale (NIHSS) [27,28], Lower Extremity Fugl Meyer (LEFM) [29,30], 6 minute walk test (6MWT) [31,32], 10 Meter Walk Test (10mWT) [33,34], Nottingham Assessment of Somato-Sensations (NSA) [35,36], Montreal Cognitive Assessment (MoCA) [37,38] and Modified Ashworth Scale of Spasticity (MAS) [39,40]. MAS ankle plantarflexion scores that were marked with a "+," an additional 0.5 points was added for calculations.

For comparison, we recruited age-matched controls who had not experienced a stroke. Exclusion criteria were: history of neurological injury, musculoskeletal damage, or current injuries that affected participants ability to move or feel either their ankles, use of medication that would change how the brain perceived pain/movement. For both stroke and age matched participants the local ethics committee approved this study, and written informed consent was obtained from each participant prior to participating, following procedures established by the University of California Irvine Institutional Review Board.

**Robotic Device**

Two versions of Ankle Measuring Proprioceptive Device (AMPD and 2AMPD, an improved version of AMPD) were used for this study. Here we briefly describe 2AMPD, which is similar to AMPD (see [41]) with some clinical usability improvements such as improved seating and robot mode switching. To use 2AMPD, participants sit in an upright position with the hip and knee bent at 90 degrees such that the shank is perpendicular to the ground, and the center of the lateral malleolus is aligned to the rotational shaft of 2AMPD. 2AMPD has two impedance states, mechanically rigid and mechanically transparent, which are chosen my manually locking or unlocking the ankle plates to/from linear screw motors (TiMOTION, TA16). In its rigid state, 2AMPD can individually assist and move both ankles via the motors through participants' natural dorsiflexion and plantar flexion passive range of motion. In its mechanically transparent state, 2AMPD allows participants to move their ankles on their own volition with minimal resistance, disconnected from the motors. 2AMPD is equipped with angular quadrature encoders to measure ankle angular position (E6B2-C, 1024 P/R) and torque cells to measure ankle torque (Interface, SMA-200). Data is acquired at 200 Hz and stored on a laptop.

**Lower Extremity Motor Impairment and Gait Function Measurements**

Standardized instructions and a demonstration were provided before each ankle impairment test, active range of motion and maximum dorsiflexion strength. For active range of motion (AROM) tests, 2AMPD was placed in its mechanically transparent state and participants were instructed to dorsiflex the ankle to their maximum position and then transition to their maximum plantarflexion position. A single trial consisted of each maximum position held for 3 seconds. Three trials were first performed on the unimpaired ankle for understanding and then the impaired ankle completed the three trials. 15 seconds of rest was given between each trial.

For maximum dorsiflexion strength, 2AMPD was placed in its mechanically rigid state with the ankle in neutral, i.e., an angle of 90° in sagittal plane between foot and shank [42]. Participants were asked to gradually dorsiflex until maximum effort was given and held for 3 seconds. The participant performed three trials with the unimpaired ankle first, and then three with the impaired ankle. 15 seconds of rest was given between each trial.

Standardized instructions and a demonstration were provided before each walk test. Each walk test was performed without participants wearing an ankle-foot orthosis. For the 10mWT, the time in seconds to walk the middle 6-meter section of a 10-meter walkway was used to compute comfortable walking speed. Timing started when the participant's first foot crossed the 2-m mark and stopped when the first foot crossed the 8-meter mark, though the participant continued to walk to the 10-meter mark [33]. Participants performed 5 repetitions, and no encouragement was given during the test.

For the 6MWT the test was performed in a corridor, and the participant was instructed to, at a comfortable pace, cover as much as ground they could during the six-minute testing period [43]. The total distance in meters was measured. Participants completed this test once, and no encouragement was given during the test.

**Crisscross Test**

For the Crisscross test, 2AMPD drove the left and right ankles in opposing directions during a series of non-periodic ankle-crossings of different angular velocities (Figure 1A). For each ankle-crossing movement, participants were instructed to press a handheld button when they perceived their feet to be at the same angular position.

For the participants who had a stroke, before beginning the test, the physical therapist assessed each participants' passive range of motion (pROM) by manually moving the unimpaired and impaired ankle with 2AMPD in its mechanically transparent mode to a comfortable maximum dorsiflexion and plantarflexion position. The feasible crossing workspace was then calculated by taking the smaller extent of dorsiflexion between the unimpaired and impaired ankle, and the smaller extent of plantar flexion as well, such that both ankles were passively moving when a crossing occurred. The assessment workspace was defined as their range of motion between maximal dorsiflexion (DF) and plantarflexion (PF), such that each joint crossing varied between >60% PF (extension), 60-20% PF (mid-extension), 20%PF - 20%DF (center), 20%-60%DF (mid-flexion) and >60% DF (flexion). This ensured a uniform distribution of crossings in the workspace. The same method was performed on the young healthy participants by a trained device operator.

To ensure each participant understood the test, they first completed four crossing movements with vision of their feet, and were asked to verbally express their understanding of the test. Then, with vision occluded by a large lap table, each participant experienced two crossing attempts in each section, to total 10 crossover movements. Participants experienced four ankle speeds: 4.4, 5.7, 7.0, and 8.3 degrees/second (Figure 1B). Individual ankle speeds were randomized such that the impaired and unimpaired ankle mostly did not move at equal speeds.

**Joint Position Reproduction (JPR) Test**

Many variations in protocols for JPR have been proposed, but the general idea is to test the ability to reproduce a passively imposed movement. Generally, JPR has shown good test–retest reliability, with an intraclass correlation coefficient between 0.60 and 0.98 and a minimal detectable change between 0.03° and 2.9° in healthy persons [44,45]. Here we implemented a passive-active contralateral JPR test, where the impaired foot was passively driven by 2AMPD and the unimpaired ankle was actively moved by the participant to try to copy the movement of the impaired ankle. Of note, what we are referring to as the "unimpaired" ankle commonly has some motor deficits resulting from the stroke as well [46,47], and thus the JPR test may be affected by these deficits.

We designed the ankle joint trajectories to have two parts, which we term the dynamic and static periods (Figure 1C). The dynamic periods consisted of 2AMPD driving the impaired ankle at a constant velocity of 5°/s through its available dorsiflexion and plantarflexion range (Figure 1C). Participants were instructed to match angular position and speed using their unimpaired ankle during these dynamic periods. Dynamic periods were randomly interrupted by static periods where the robot stopped moving the ankle at a pseudo-random set of positions distribution across the workspace. Participants were instructed to match the angular position of the stationary impaired ankle by making fine adjustments with their unimpaired ankle. For each static period, unlimited time was given, and participants were instructed to press a handheld button when they perceived their feet to be at the same angular position. The robot returned to a dynamic period after they pushed the button. Just like for Crisscross, participants first performed a short practice test with vision of their feet allowed and were asked to verbally express their understanding of the test. Then, their vision of their feet was occluded with the lap table and the subsequent test lasted about 2 minutes.

Since Crisscross was completed before the JPR test, the pROM of the impaired ankle used in Crisscross was used in the JPR test and 2AMPD drove the impaired ankle to 80% of the maximum dorsiflexion and plantarflexion positions. The static periods were selected by splitting the impaired ankle ROM into four regions, >60% PF (extension), 60-10% PF (mid-extension), 10%PF - 60%DF (mid-flexion), >60%DF (flexion). One static period occurred in each region resulting in a total of 4 dynamic periods and 4 static periods. The same method was performed on the age matched controls, but 2AMPD passively drove the nondominant ankle and participants used the dominant ankle as the indicator. For age matched healthy controls dominant ankle was determined by asking participants which foot they used to kick a ball with.

FIGURE 1

## Statistical Analysis

**Ankle Proprioceptive Acuity**

Proprioceptive acuity using Crisscross was quantified using absolute error, defined as the absolute angular difference between the left and right ankle at the moment of button press. If a participant did not attempt to press the button on single or multiple crossing attempts, their average error was calculated using only the crossings where a button press happened. Similarly, proprioceptive acuity using JPR was also quantified using absolute error. Absolute error here was defined as the absolute angular difference between the left and right ankle at the moment of button press for the static condition, and the absolute angular difference averaged across the dynamic condition.

**Ankle Motor Impairment**

For AROM and maximum dorsiflexion strength tests, only the impaired ankle was considered. The maximum dorsiflexion and plantarflexion position of the 3 trials were averaged. Then the maximum dorsiflexion and plantarflexion position was summed together for a total active range value. For maximum strength, the maximum torque produced in each trial was taken and averaged across all three trials for the impaired ankle.

**Gait Function**

The 5 repetitions for the 10mWT were first converted to a velocity (meters/sec), and then averaged. For the 6MWT, the total distance was used in all gait function analysis. If stroke participants were unable to complete the 10mWT or 6MWT without an AFO, they were given zero for each test they could not complete.

Statistical analyses were conducted using Matlab R2023 software. Each output parameter was independently tested for normality using Shapiro Wilks test. The 10mWT and maximum dorsiflexion strength were not normally distributed ($p < 0.05$), but 6mWT, active range of motion, and LEFM were normally distributed ($p > 0.05$). Since not all data series proved to be normally distributed, we used non-parametric tests, Wilcoxon rank-sum test for comparison and Spearmans's rank order for correlation. An alpha level of 0.05 was used for all comparisons and correlations.

# Results

**Participants**

Thirty-nine people in the chronic phase of stroke participated in the study (mean age 59 ± 12 SD; 15 female/24 male). 27 were left side affected and 12 were right side affected. Table 2 provides a demographic and clinical overview of the participants. Sixteen non-impaired, age-matched controls were included (65 ±11yrs; 7F/9M, $p = 0.15$). 12 participants were right-side dominant, and 4 participants were left-side dominant.

TABLE 2

**Overview of Proprioception Assessment Results**

For Crisscross, stroke participants pushed the button on 333 out of the 390 crosses. 23 participants attempted all 10 crosses that were presented. 16 participants missed at least 1 crossing, and, of these, 4 pushed the button on less than 50% of crossings. Of the 57 total crosses with no button press, 29 attempts were missed in the plantarflexion region and the remaining in the dorsiflexion region. Of the 333 crossings attempted, participants pressed the button on average 0.3 ± 1.3 seconds before crossing indicating that they on average overly anticipated the moment their ankles would cross. The average absolute error for the Crisscross test was 16.3°±7.2°.

For JPR, during the dynamic period participants lagged the target on average 50%±15% of the time and led 50%±15% of the time. The average speed at which participants moved the unimpaired ankle during the dynamic phase was 5.1°/s ± 2.0°/s, which was not significantly different from the average speed of the impaired ankle that they were trying to track as the robot moved it (5.0°/s ±0.2°/s, $p = 0.99$). Of the 156 static periods, 59% of button presses occurred with the unimpaired ankle below the intended position. Participants on average pressed the button 5.6 ± 8.0 seconds after the start of the matching period. The delay to button press was positively correlated with JPR static error ($\rho = 0.32$, $p = 0.046$); thus, participants who took longer to press the button to indicate they had matched their ankle positions exhibited greater error. The average absolute error for JPR Static and JPR Dynamic was 11.3°±7.1° and 11.2°±5.6°, respectively.

For age matched participants their average Crisscross error, JPR Static error, and JPR Dynamic error were: 9.6° ± 3.5°, 6.1° ± 2.3°, 6.2° ± 2.0°, respectively. Impaired proprioception, defined as exhibiting a mean error that was greater than 2 SDs of the mean error in healthy controls, was present in the following percentage of the participants with stroke: 87% for Crisscross, 74% for JPR Static, 76% for JPR Dynamic.

**Relationship between Robotic Assessments of Ankle Proprioception and Gait Speed**

Together, the two robotic proprioceptive assessments produced three measures of proprioceptive error: Crisscross error, JPR dynamic error, and JPR static error. The two clinical assessments of gait function produced two measures of gait speed, one based on 10 meters of walking (i.e. the 10mWT) and one based on six minutes of walking (i.e. the 6MWT). The average gait velocity for the 10mWT and meters walked for 6MWT was $0.37 \pm 0.24$ meters per second and $113.3 \pm 66.8$ meters, respectively. Table 1 shows the average values for each measure, while Figure 2 shows graphs of the measures of ankle proprioception versus the measures of gait speed. The only significant correlation was between JPR Dynamic proprioceptive error and 6MWT (JPR Static: $\rho = -0.34$, $p = 0.039$, Figure 2F). JPR Dynamic error was nearly significantly related to gait speed in the 10mWT ($\rho = -0.28$, $p = 0.09$, Figure 2C) and Crisscross error was nearly significantly related to 6MWT distance ($\rho = -0.29$, $p = 0.08$, Figure 2D).

FIGURE 2

**Relationship between Robotic Assessment of Ankle Proprioception and Lower Extremity Motor Impairment**

The average AROM for the impaired ankle was 31.7° ± 16.7°. The average maximum dorsiflexion strength was 9.7 ± 7.4 Nm. The average LEFM score was 20 ± 3 (out of a possible 34, Table 1). Figure 3 shows graphs relating ankle proprioception error to measures of ankle motor impairment. No significant relationships were found between ankle proprioception and ankle motor impairment ($p > 0.2$).

FIGURE 3

## Discussion

Gait function is a key factor in determining the level of independent mobility during activities of daily living after stroke [48]. Identifying the specific motor and sensory impairments that influence gait function is of great interest in stroke rehabilitation research, in part because this knowledge helps guide treatment. In this study, we quantified ankle proprioceptive ability of 39 individuals in the chronic phase of stroke. Using a novel robotic device we implemented two bilateral proprioceptive tests, a joint position reproduction (JPR) test similar to previously developed JPR tests, and the Crisscross test, which here we applied to the measurement of ankle proprioception after stroke for the first time. Using a 2SD criteria relative to age-matched controls, we found that ankle proprioception deficits were common in our participants, being present in approximately 75-90% of individuals we tested, depending on the specific test and error measure. We investigated the relationships between the magnitude of the proprioceptive error measured with these tests and gait function, quantified as gait speed across long and short distances. We found only one significant but weak relationship between ankle proprioceptive acuity (measured with JPR dynamic error) and gait function (measured with the 6MWT). We also found that ankle proprioception acuity was not significantly related to three measures of LE motor impairment (ankle AROM, ankle dorsiflexion strength, and LEFM score). We discuss first the significance of these results then limitations and directions for future research.

**Relationship between Ankle Proprioception and Gait Function**

After stroke, significant but weak associations between LE sensation and gait impairment have sometimes been observed [7,8,19–21] but not always [5–8] (see Table 1). This is somewhat surprising as there is a large body of research that has identified the importance of LE sensory input for locomotion plasticity [15–18]. It has been suggested that the inconsistency in findings may be explained by the variation in methods used to quantify proprioception [7,19]. Here we applied two robotic ankle proprioception assessments methods, hoping that the improved consistency and sensitivity provided by robotics might more definitively answer this question. Yet we still found a mixture of significant, moderately insignificant, or insignificant correlation results depending on the test, the measure of proprioceptive error, and the measure of gait function; in all cases the putative correlations were weak in magnitude. This result would be explained if: 1) there is only a weak relationship between proprioception and gait function after stroke; and 2) there is high variability in proprioception acuity between individuals after stroke. Then, regardless of the sensitivity of the proprioception test, one would expect to find weak correlations, and that the statistical significance would depend on the particular sample of participants.

Why might the relationship between ankle proprioception and gait function after stroke be weak? One possibility is that, while normal gait function relies on ankle proprioception, individuals who lose ankle proprioception learn to compensate using other sensory pathways. Ankle proprioception is thought to rely mostly on information from muscle spindles [49], but if spindle pathways are damaged then the locomotor control system could substitute information from cutaneous and load related afferents, which are modulated during gait due to loading and weight-shifting [50]. Furthermore, the central nervous system (CNS) may also reduce reliance on somatosensory information and increase reliance of visual and vestibular inputs [7]. The relative contribution of somatosensory, visual, and vestibular sensory inputs changes in response to individual, task, and environmental factors [51,52]. Measuring gait function in the dark after stroke might reveal a greater dependence on loss of integrity of leg proprioception.

Why is there high variability in ankle proprioceptive function after stroke? The neuroanatomical damage due to stroke is highly variable in its location and extent; somatosensory structures are sometimes affected and at other times spared. Further, ankle proprioception testing relies on cognitive abilities such as attention and working memory, which are commonly impaired post-stroke [53] and confounded by fatigue [54]. These impairments may further increase proprioceptive testing variability.

**Relationship between Ankle Proprioception and Lower Extremity Motor Impairment**

Consistent with other studies [5–8] we did not find a significant relationship between ankle proprioception and LE motor impairment in our sample of persons with a stroke. No relationship we tested had a significance value less than 0.2. It might be that other measures of LE motor impairment might lead to significant results, but the three we tested here – active ROM, dorsiflexion strength, and LEFM score – are widely used and clinically relevant. A more likely possibility is that the neural tracts and circuits supporting ankle proprioception are anatomically distinct from those supporting LE motor function. Thus, when a stroke destroys ankle proprioceptive circuitry, it is unlikely to damage LE motor circuitry in a proportional way. Although the LE motor and somatosensory representations in primary motor and somatosensory cortex neighbor each other [55,56], sensory-motor control of LE motion during walking is to some degree offloaded to the spinal cord [57–59]. Thus, damage to cortical sensory-motor areas may induce LE sensory deficits but have smaller consequences for gait function.

**Limitations**

This study has several limitations. First, while the sample size of 39 is considered adequate in statistical theory for performing correlation analysis [60], increasing the sample size might make the detection of any weak correlative relationships more robust. Second, we deployed only two specific proprioceptive tests, and only measured proprioceptive acuity at the ankle joint. A recent review by Horvath et al., highlights that errors measured with different proprioceptive assessments or at different joints do not appear to generalize, at least for young unimpaired persons and persons with peripheral nervous system damage, suggesting that each proprioceptive assessment tests a different aspect of proprioception [23]. If different proprioceptive assessments test different underlying mechanisms, changing the method of proprioceptive assessment may change the result. Third, there may be aspects of gait function with which ankle proprioception is more strongly related. For example, this study did not test walking in dark conditions or include an assessment specifically focused on balance [16], although we would expect deficits in balance should be reflected in gait speed.

# Acknowledgements

This study was supported by an operating grant from the National Institute of Health (5R01HD095457and NIH R01HD062744) and NIDILRR (90REGE0010).

**Table 1:** Summary of studies of the relationship between LE sensation and gait function after stroke.

| Study (# of Participants) | Measure of Lower Extremity Sensation | Measure of Gait Function | Correlation Result |
|---|---|---|---|
| Nadeau et al. 1999 (16) | LEFM | 9mWT | r = 0.14 |
| Hsu et al. 2003 (26) | LEFM | 6mWT | ρ = 0.4* |
| Lee et al. 2005 (11) | TDPM (robot) | 6MWT | ρ = 0.63 to 0.77* |
| Lin 2005 (21) | JPR (robot) | 6mWT | r = -0.021 |
| Gorst et al. 2018 (32) | Gradient Dscr and Step Height Dscr (robot) | 10mWT<br>Balance – COP | r = -0.40 to -0.60*<br>r = -0.43 to -0.44* |
| Gorst et al. 2019 (163) | EmNSA | Falls Efficacy Scale<br>Balance - Centre of force<br>10mWT | ρ = -0.22*<br>ρ = -0.20*<br>ρ = 0.09 |
| Cho et al. 2021 (57) | JPR (robot) | Berg Balance Scale | ρ = -0.40* |

**Abbreviations: LEFM** = Lower Extremity Fugl Meyer; **TDPM** = Threshold to Detection of Passive Motion; **JPR** = Joint Position Reproduction; **Dscr** = Discrimination; **EmNSA** = Erasmus MC modified version of the Nottingham Sensory Assessment; **COP** = Center of Pressure; **ρ** = Spearman's rank correlation; **r** = Pearson's correlation; ***** denotes significant relationship ($p < 0.05$)

ankle proprioception and gait control

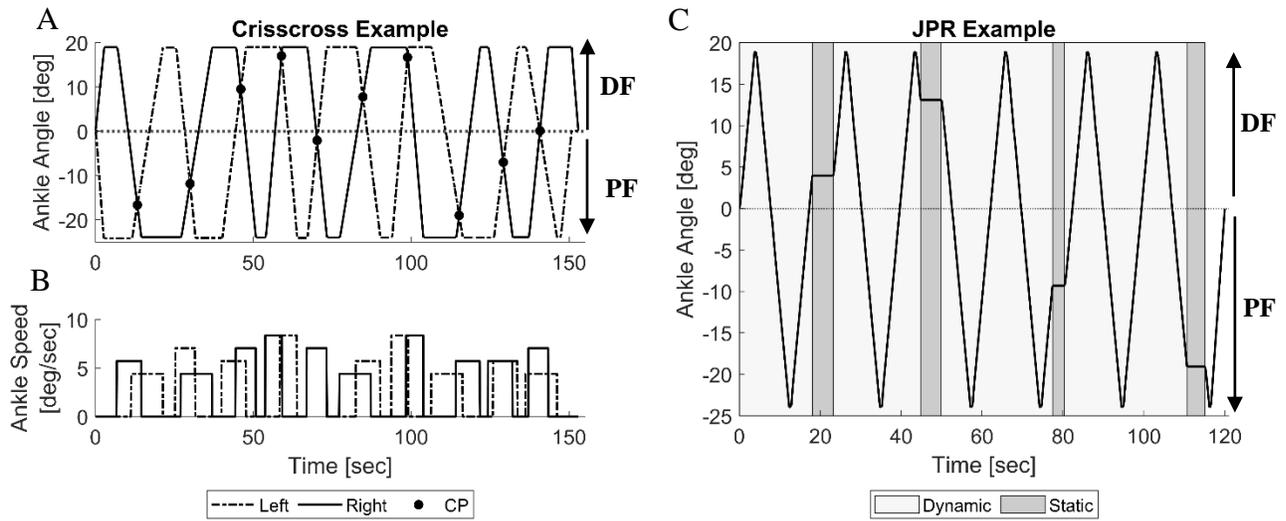

**Figure 1**. Examples of Crisscross and JPR tests. **A)** Ankle trajectories and crossing positions (denoted by circles) during Crisscross test generated by 2AMPD. Positive ankle angle corresponds to dorsiflexion. **B)** Ankle speeds during Crisscross test generated by 2AMPD. **C)** Unimpaired ankle trajectory during JPR test generated by 2AMPD, showing both dynamic and static periods.
**DF** = Dorsiflexion Direction; **PF** = Plantarflexion Direction

**Table 2:** Characteristics of the participants with a stroke

|  | **Average ± SD** | **[Min Max]** |
|---|---|---|
| Age | 60 ± 12 | [27 76] |
| Days Post Stroke | 1155 ± 1096 | [201 4085] |
| NIH Stroke Severity Scale [0 42] | 6 ± 3 | [2 16] |
| Lower Extremity Fugl Meyer [0 34] | 20 ± 3 | [12 26] |
| Modified Ashworth Score [0 4] | 1.59 ± 0.48 | [0 2] |
| 6 min walk distance (meters) | 113.3 ± 66.8 | [0.20 298.50] |
| 10MWT (m/s) | 0.37 ± 0.24 | [0 0.76] |
| Montreal Cognitive Assessment (MoCA) [0 30] | 23 ± 6 | [1 29] |
| Ischemic/Hemorrhagic/Both | 20/16/3 | |
| Biological Sex M/F | 24/15 | |

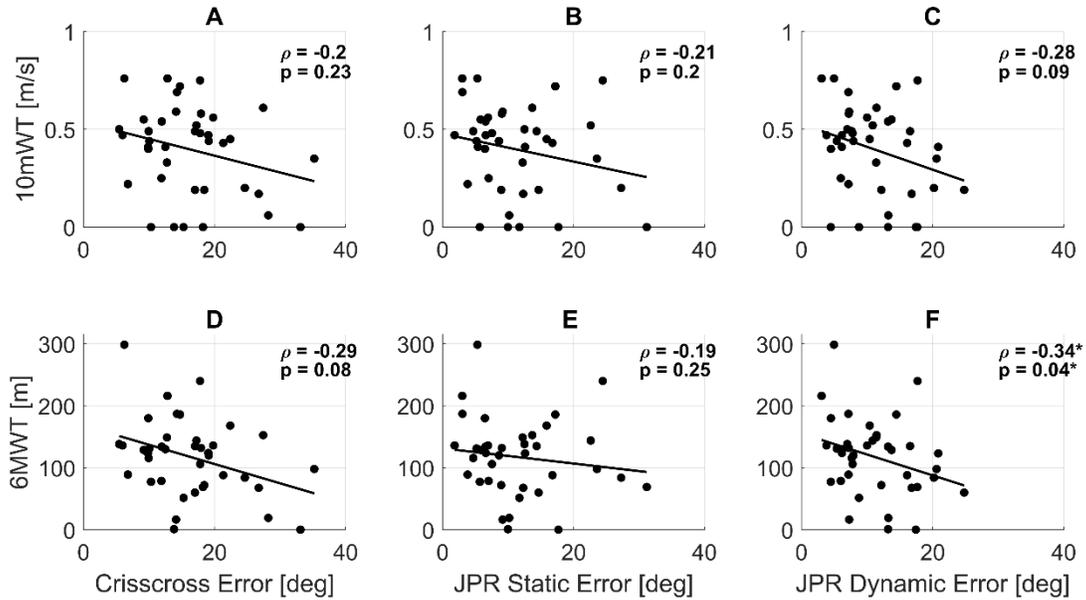

**Figure 2**. Relationship between three robotic measures of ankle proprioception (x axis) and two measures of measures and gait function (y axis). **A, B, C:** Average 10mWT gait velocity versus proprioceptive error. **D, E, F:** 6MWT distance versus proprioceptive error. Statistics from applying Spearman's correlation are shown.
**Abbreviations:** *10mWT*: 10-meter walk test; *6MWT:* 6-minute walk test; * denotes p < 0.05

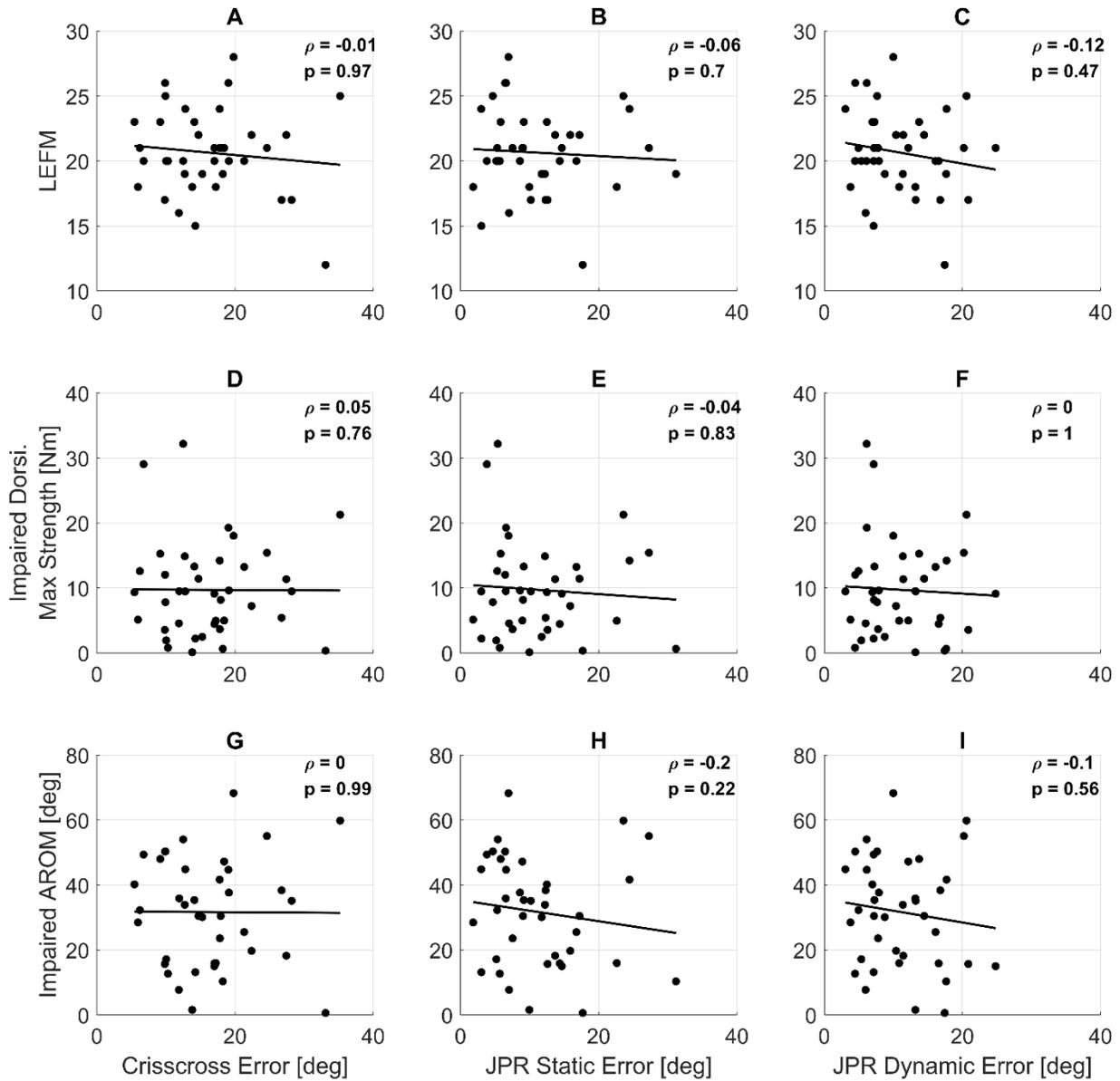

**Figure 3**. Relationship between three robotic measures of ankle proprioception (x axis) and three measures of measures of LE motor impairment (y axis) **D, E, F:** Impaired dorsiflexion maximum strength as a function of proprioceptive error. **G, H, I:** Impaired AROM as a function of proprioceptive error.
**Abbreviation:** LEFM: Lower Extremity Fugl Meyer; AROM: Active Range of Motion; * p < 0.05